\documentclass[superscriptaddress, groupedaddress, notitlepage, amsmath, amssymb, showkeys, aps, pre]{revtex4-2}

\usepackage{graphicx}

\usepackage{hyperref}

\usepackage[mathlines]{lineno}

\usepackage{gensymb}
\usepackage{physics}
\usepackage{mhchem}

\usepackage[text={7.in,9.5in},centering]{geometry}

\usepackage{tabularray}
\usepackage{enumitem}
\usepackage{dcolumn} 

\usepackage{tikz}
\usetikzlibrary{arrows.meta, positioning, shapes.geometric, fit, backgrounds, calc}

\usepackage{orcidlink}

\begin{document}

\title{Stan: An LLM-based thermodynamics course assistant}

\author{Eric M.\ Furst\,\orcidlink{0000-0002-8849-4484}}
\email{furst@udel.edu}
\affiliation{Department of Chemical and Biomolecular Engineering, Allan P.\ Colburn Laboratory, University of Delaware, Newark, DE 19716, USA}

\author{Vasudevan Venkateshwaran\,\orcidlink{0000-0002-0690-0056}}
\email{vvenkate@udel.edu}
\altaffiliation{Also W. L. Gore \& Associates Inc., 100 Airport Road, Elkton, MD 21921, USA}
\affiliation{Department of Chemical and Biomolecular Engineering, Allan P.\ Colburn Laboratory, University of Delaware, Newark, DE 19716, USA}

\date{\today}

\begin{abstract}
Discussions of AI in education focus predominantly on student-facing tools---chatbots, tutors, and problem generators---while the potential for the same infrastructure to support instructors remains largely unexplored.  We describe Stan, a suite of tools for an undergraduate chemical engineering thermodynamics course built on a data pipeline that we develop and deploy in dual roles: serving students and supporting instructors from a shared foundation of lecture transcripts and a structured textbook index. On the student side, a retrieval-augmented generation (RAG) pipeline answers natural-language queries by extracting technical terms, matching them against the textbook index, and synthesizing grounded responses with specific chapter and page references.  On the instructor side, the same transcript corpus is processed through structured analysis pipelines that produce per-lecture summaries, identify student questions and moments of confusion, and catalog the anecdotes and analogies used to motivate difficult material---providing a searchable, semester-scale record of teaching that supports course reflection, reminders, and improvement. All components, including speech-to-text transcription, structured content extraction, and interactive query answering, run entirely on locally controlled hardware using open-weight models (Whisper large-v3, Llama~3.1 8B) with no dependence on cloud APIs, ensuring predictable costs, full data privacy, and reproducibility independent of third-party services.  We describe the design, implementation, and practical failure modes encountered when deploying 7--8 billion parameter models for structured extraction over long lecture transcripts, including context truncation, bimodal output distributions, and schema drift, along with the mitigations that resolved them.
\end{abstract}


\maketitle

\section{Introduction and context}

We present Stan---a suite of learning tools designed to support instruction and student engagement in an undergraduate chemical engineering thermodynamics course. Stan broadly consists of a Large Language Model (LLM)-based semantic query tool, a lecture transcriber aligned with the course textbook \cite{sandler2017chemical} and lecture material, and a command-line based interface for students and instructors to run queries. 

The prevailing discussion of AI in the classroom focuses on \emph{student-facing} tools: chatbots that answer questions, tutors that explain concepts, and generators that produce practice problems. \cite{uiuc_ai_schools_2024, thomson2024impact, klimova2025effects, mcgehee2025aistudent, palmer2025aistudent,kasneci2023,lo2023,yan2024,denny2024a,mollick2023,loubet2026} We argue that an equally important and under-explored dimension is the set of \emph{instructor-facing} tools that help faculty manage and analyze their own course and teaching. A searchable index over lecture transcripts, for example, serves students who want to find when a topic was covered, but it also allows the instructor to identify recurring questions, pinpoint areas of persistent confusion, trace how the treatment of a concept evolved across the semester, or even catalog the anecdotes and analogies they used to motivate difficult material. 

This dual perspective shapes the design of Stan. On the student side, the system functions as an interactive, context-aware resource that complements classroom instruction. Students pose natural language queries and receive coherent, context-sensitive responses grounded in the textbook and lecture recordings. On the instructor side, the same underlying data and retrieval infrastructure supports course analytics---aggregating student queries to surface common misconceptions, summarizing lecture content for review, and providing a feedback channel that operates outside of office hours and formal course evaluations. 

Building the system requires several technical challenges to be addressed. The most important is that LLM-generated responses are subject to hallucinations---confident but incorrect outputs---which can be particularly problematic in precision-focused subjects like thermodynamics. \cite{huang2025} Computational capacity, student training, security and privacy risks (e.g.\ prompt injection and data leakage) are other considerations that need to be addressed when deploying  such systems in academic settings. \cite{yan2024}

One method to address these concerns is to run open models on local hardware rather than proprietary models using cloud-based APIs that provide limited visibility to how the risks outlined above are handled. \cite{kolagar2024} 
There is another advantage in this approach---the open and relatively easy to understand tools are Python scripts and are modifiable by students and instructors alike to be tailored and enhanced with new features. With additional data, such as their own notes from this course (as well as other courses in the curriculum) students will be able to run more complex queries. For example, one can imagine supplementing notes from a linear algebra or numerical methods course and get context relevant information on how mathematical methods from there apply to solving phase equilibria problems. 

In this article we describe in detail the design and implementation of all components of Stan, providing additional context and rationale for our design choices. We begin by outlining the goals of Stan and the design decisions that shaped its development, then describe the hardware and deployment architecture required to run it. We then walk through the processing pipeline for the lecture audio transcription tool and topics index, which serve as the primary data sources against which queries are run. We present the two primary interfaces built for Stan---a student-facing textbook query tool and an instructor-facing lecture analysis tool---along with example queries for each. Finally, we discuss directions for future work and potential extensions of the system, including enabling Stan with the capability to run interactive code examples.

\section{Goals of Stan}

Stan's goals are organized along two axes: the \emph{capability level} of the learning assistant (what it can do) and the \emph{audience} it serves (students or instructor).

\subsection{Student-facing capabilities}

For students, we define a tiered task-based capability scale for the learning assistant. This classification is not a formal taxonomy but is intended to be understandable and practical; it is inspired by capability taxonomies such as Bloom's. \cite{zaphir2024critically} At this point in time, our goal for Stan is to build a system operating at the level of a Resource Pointer (Level 1) and Content Summarizer (Level 2) as defined in Table~\ref{tab_learning_levels}---a tool that not only directs students toward relevant textbook sections and course materials, but also connects lecture content to those materials---functioning as a course scribe that captures and recycles the information presented in class.

\begin{table}
\caption{\label{tab_learning_levels}%
Classification of Learning Assistant Intelligence Levels}
\begin{tblr}{hlines, colspec={lllQ[p,wd=0.4\textwidth]}}     
\hline
\textbf{Level} & \textbf{Title} & \textbf{Primary Role} & \textbf{Capabilities} \\
1 & Resource Pointer & Indexer & 
	\begin{itemize}[leftmargin=*, noitemsep, topsep=0pt]
	\vspace*{-\baselineskip}
 		\item Points to textbook sections or course materials 
 		\item Acts as a smart search tool without explanations 
 	\end{itemize} \\
 2 & Content Summarizer & Summarizer & 
 	\begin{itemize}[leftmargin=*, noitemsep, topsep=0pt]
	\vspace*{-\baselineskip}
	\item Summarizes textbooks, lectures, or documents 
	\item Produces concise, digestible content
	\end{itemize} \\
3 & Concept Explainer & Tutor &
	\begin{itemize}[leftmargin=*, noitemsep, topsep=0pt]
	\vspace*{-\baselineskip}
	\item Explains concepts in plain language; Uses examples, analogies, and context
	\item Answers why and how questions
	\end{itemize} \\

4 & Guided Problem Solver & Step-by-step Instructor & 
	\begin{itemize}[leftmargin=*, noitemsep, topsep=0pt]
	\vspace*{-\baselineskip}
	\item Solves problems step-by-step; Provides hints and guiding questions
	\item Identifies common mistakes 
\end{itemize} \\
5 & Exercise Generator & Practice Designer & 
	\begin{itemize}[leftmargin=*, noitemsep, topsep=0pt]
	\vspace*{-\baselineskip}
	\item Suggests custom exercises and practice problems 
	\item Tailors questions to the topic without tracking performance 
\end{itemize}\\
6 & Autonomous Learning Companion & AI Collaborator & 
	\begin{itemize}[leftmargin=*, noitemsep, topsep=0pt]
	\vspace*{-\baselineskip}
	\item Engages in Socratic-style dialogue; Supports metacognitive thinking
	\item Integrates cross-subject concepts \end{itemize} \\
\hline
\end{tblr}  
\end{table}

To enable this we use a hierarchical index of topics developed from the textbook, and transcripts of the lecture audio as the primary data sources. A detailed description of the data processing pipeline and the required hardware are presented in Section \ref{sec:design_and_tech_stacks}.

\subsection{Instructor-facing capabilities}
\label{sec:instructor_facing}

We believe that instructor-facing tools represent an equally valuable and largely untapped application of the same underlying infrastructure. The datasets and tools underlying Stan's student facing capabilities also  provides the instructor with a searchable record of their own teaching. The same index and pipeline that answers ``Where was entropy generation discussed?'' for a student can answer ``What questions did students struggle with most in the last three lectures?'' for the instructor. The difference lies not in the technology but in the query and the richness of the dataset.

Concrete use cases for instructors include:
\begin{itemize}
\item \textbf{Content reinforcement.} The system can identify material that was covered verbally in lecture but never appeared in written form (homework,   slides, textbook reading). This helps the instructor close gaps between what was said and what students have to study from.
\item \textbf{Lecture analytics.} Searching the transcript corpus allows the instructor to trace how concepts were introduced and developed across the semester: ``When did I first discuss fugacity?'', ``How much lecture time did I devote to departure functions?'', ``Did I connect the Clausius-Clapeyron equation back to the phase diagrams from Chapter 5?''
\item \textbf{Course evolution.} Comparing transcripts across semesters reveals how the treatment of topics changes over time---an institutional memory that is otherwise lost when lectures are not recorded or indexed. We envision processes that help coordinate between instructors or provide new tools for the department curriculum committee.
\item \textbf{Question mining.} By logging and analyzing the queries that students pose to the system, the instructor can identify recurring questions, common misconceptions, and topics that students find most difficult---without relying solely on office hours or end-of-semester evaluations. This creates a passive, continuous feedback channel that captures student confusion in real time throughout the course. 
\end{itemize}

\section{Design and technical stack}
\label{sec:design_and_tech_stacks}

A deliberate design choice in Stan is that all components run on locally controlled hardware, with no dependence on cloud APIs or external services. This provides several advantages in an academic setting: predictable costs (no per-query API fees), full control over student data and privacy, the ability to operate without internet access, and reproducibility independent of third-party service changes.

\subsection{Hardware and deployment architecture}

The system uses two hardware tiers:
\begin{itemize}
\item \textbf{GPU workstation.} A Linux workstation equipped with a single NVIDIA RTX 4090 (24\,GB VRAM) handles compute-intensive tasks: Whisper transcription (large-v3, float16) and, when needed, local LLM inference for larger models. At current prices, this represents a one-time hardware investment comparable to a single year of cloud API costs for equivalent throughput.
\item \textbf{Consumer laptop.} The RAG retrieval pipeline, vector search, and interactive query interface run on a standard consumer laptop using Ollama for local LLM inference. \cite{ollama2023} Models in the 7--13\,B parameter range (e.g.\ Llama, Mistral, Qwen, Apertus) run comfortably on laptop-class hardware with or without a discrete GPU, making the student-facing components deployable on hardware that instructors and students already own. 
\end{itemize}

This two-tier architecture separates the batch processing pipeline (transcription, index building) from the interactive query pipeline (retrieval, LLM synthesis). The GPU workstation performs the heavy lifting once---transcribing a semester of lectures in under 45 minutes---and produces artifacts (text transcripts, vector indices) that are lightweight enough to distribute and query on any machine.

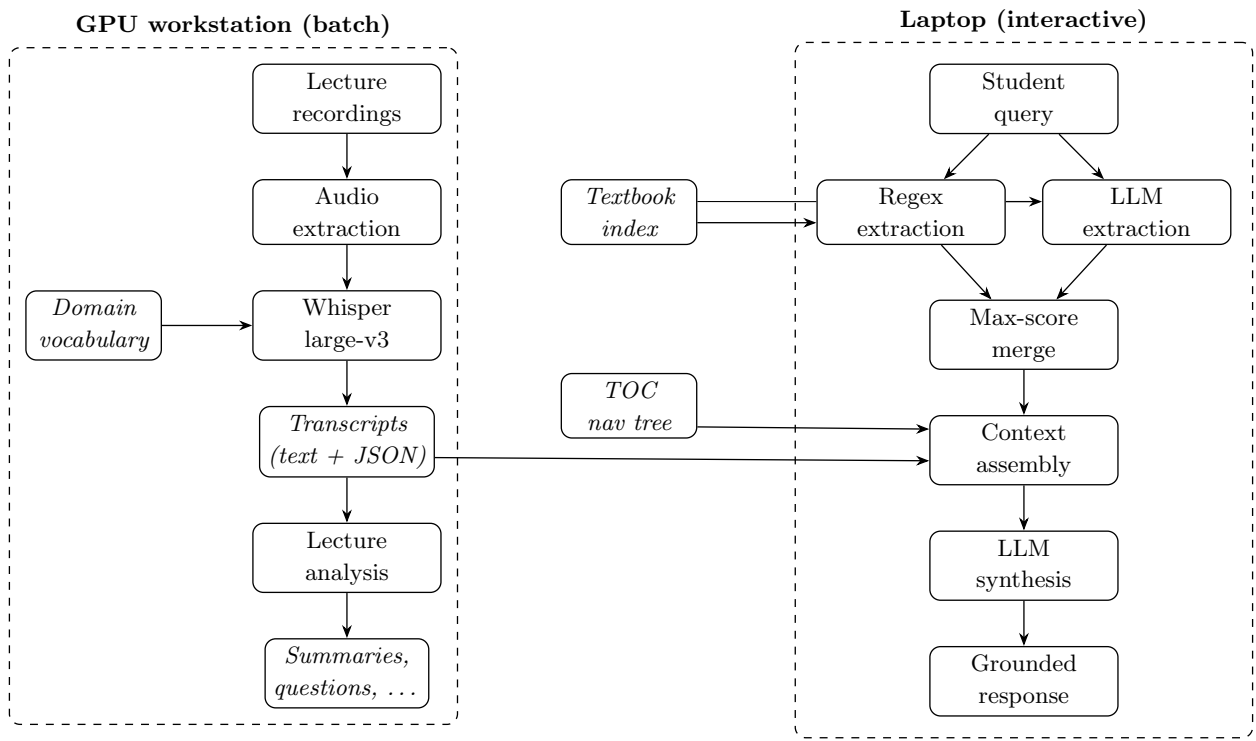
\begin{figure*}
\centering
\begin{tikzpicture}[
  node distance=0.6cm and 1.0cm,
  box/.style={draw, rounded corners, minimum width=2.5cm,
              minimum height=0.7cm, align=center, font=\small,
              line width=0.5pt},
  data/.style={draw, rounded corners, minimum width=1.8cm,
               minimum height=0.6cm, align=center, font=\small\itshape,
               line width=0.5pt},
  hw/.style={draw, dashed, rounded corners, inner sep=6pt,
             font=\small\bfseries, line width=0.5pt},
  arr/.style={-{Stealth[length=5pt]}, line width=0.5pt},
]
\node[box] (capture) {Lecture\\recordings};
\node[box, below=of capture] (audio) {Audio\\extraction};
\node[box, below=of audio] (transcribe) {Whisper\\large-v3};
\node[data, below=of transcribe] (transcripts) {Transcripts\\(text + JSON)};
\node[box, below=of transcripts] (analyze) {Lecture\\analysis};
\node[data, below=of analyze] (analysis) {Summaries,\\questions, \ldots};
\draw[arr] (capture) -- (audio);
\draw[arr] (audio) -- (transcribe);
\draw[arr] (transcribe) -- (transcripts);
\draw[arr] (transcripts) -- (analyze);
\draw[arr] (analyze) -- (analysis);
%
\node[data, left=1.2cm of transcribe] (vocab) {Domain\\vocabulary};
\draw[arr] (vocab) -- (transcribe);
%
\begin{scope}[on background layer]
  \node[hw, fit=(capture)(audio)(transcribe)(transcripts)(analyze)(analysis)(vocab),
        label={[font=\small\bfseries]above:GPU workstation (batch)}] (gpubox) {};
\end{scope}
%
\coordinate (rorigin) at ($(capture.north) + (9.0cm, 0)$);
\node[box, anchor=north] at (rorigin) (query) {Student\\query};
\node[box, below=0.6cm of query, xshift=-1.5cm] (regex) {Regex\\extraction};
\node[box, below=0.6cm of query, xshift=1.5cm] (llm_ext) {LLM\\extraction};
\node[box, below=2.2cm of query] (merge) {Max-score\\merge};
\node[box, below=of merge] (context) {Context\\assembly};
\node[box, below=of context] (synth) {LLM\\synthesis};
\node[box, below=of synth] (answer) {Grounded\\response};
\draw[arr] (query) -- (regex);
\draw[arr] (query) -- (llm_ext);
\draw[arr] (regex) -- (merge);
\draw[arr] (llm_ext) -- (merge);
\draw[arr] (merge) -- (context);
\draw[arr] (context) -- (synth);
\draw[arr] (synth) -- (answer);
%
\begin{scope}[on background layer]
  \node[hw, fit=(query)(regex)(llm_ext)(merge)(context)(synth)(answer),
        inner sep=8pt,
        label={[font=\small\bfseries]above:Laptop (interactive)}] (lapbox) {};
\end{scope}
%
\node[data] at ($(audio)!0.5!(regex)$) (index) {Textbook\\index};
\node[data, below=1.7cm of index] (toc) {TOC\\nav tree};
%
\draw[arr] ([yshift=-4pt]index.east) -- ([yshift=-4pt]regex.west);
\draw ([yshift=+4pt]index.east) -- ([yshift=+4pt]regex.west);
\draw[arr] ([yshift=+4pt]regex.east) -- ([yshift=+4pt]llm_ext.west);
\draw[arr] ([yshift=-8pt]toc.east) -- ([yshift=+8pt]context.west);
\draw[arr] ([yshift=-6pt]transcripts.east) -- ([yshift=-4pt]context.west);
\end{tikzpicture}
\caption{\label{fig:pipeline}%
Stan system architecture. The batch pipeline (left) runs on a GPU
workstation, producing transcripts and structured lecture analyses. The
interactive query pipeline (right) runs on a consumer laptop, using a
dual-path extraction strategy (regex and LLM) to match student queries
against the textbook index, then synthesizing grounded responses with
chapter and lecture context.}
\end{figure*}

Figure~\ref{fig:pipeline} shows the overall data flow.  Lecture recordings enter the batch pipeline (left), which produces transcripts and structured analyses that support both tools. The interactive query pipeline (right) shows the student-facing tool, where queries are matched against the textbook index and lecture corpus to generate grounded responses. The instructor-facing lecture analysis tool consumes the same transcripts through a separate analysis pipeline.

\subsection{Data sources}

\subsubsection{Textbook representation}
The retrieval pipeline draws on two structured representations of the textbook, both extracted automatically from the PDF:

\begin{itemize}
\item \textbf{Textbook index} (\texttt{bindex\_tab.json}). A hierarchical JSON structure containing approximately 1{,}500~index entries with topics, page numbers, and nested subtopics. For example, \emph{Acidity of solutions} contains subtopics for \emph{buffer}, \emph{Henderson-Hasselbalch equation}, and \emph{strong acid with strong base}, each with associated page references. This curated index  represents the textbook author's own organization of concepts. 
\item \textbf{Table of contents navigation tree} (\texttt{ftoc\_nav\_tree.json}). A structured representation of the textbook's chapter and section hierarchy, mapping page numbers to chapter titles and section headings. This allows the system to enrich raw page-number references with structural context (e.g., ``page~102 is in Chapter~4, Section~4.1: Entropy: A New Concept'').
\end{itemize}

We use a structured index search rather than vector similarity because the textbook's back-of-book index is itself a high-quality retrieval artifact. It represents the textbook author's expert curation of which concepts appear where. Embedding-based retrieval over raw textbook text would require chunking decisions, an embedding model, a vector store, and careful tuning of similarity thresholds---added complexity without clear benefit when a professionally curated index is available.

The use of the textbook's back-of-book does raise a copyright consideration, however. While the index consists of factual mappings (topics to page numbers), the selection and arrangement of entries represents the author's and publisher's editorial work and may be protected. For development and testing, this use likely qualifies as fair use under U.S.\ copyright law. It is non-commercial, educational, and the index is a small fraction of the work. However, deploying the extracted index directly to students could be viewed differently. For production deployment, we recommend either seeking explicit permission from the publisher, or replacing the published index with an instructor-generated topic list derived from the course syllabus and lecture notes---an approach that also has the advantage of being tailored to the specific course offering rather than the full textbook scope.

\subsubsection{Lecture transcripts}
Both the student-facing and instructor-facing tools depend on accurate text transcripts of lecture recordings. This section describes the automated pipeline for converting video recordings from the Fall~2025 offering of CHEG231 (Chemical 
Engineering Thermodynamics I) at the University of Delaware into searchable text.

\paragraph{Audio processing and transcription pipeline.}
Lecture recordings were captured via the university's UDCapture system (Kaltura) and downloaded in MP4 format. Audio was extracted using \texttt{ffmpeg} at 16\,kHz mono (the native input rate of the Whisper model family), then transcribed with the \texttt{faster-whisper} library, \cite{fasterwhisper2024} which provides CTranslate2-optimized inference for Whisper models. \cite{radford2023robust}

All transcriptions used the \texttt{large-v3} model in float16 precision on an NVIDIA RTX 4090 GPU. Voice Activity Detection (VAD) via Silero-VAD \cite{silero2021vad} was enabled to skip silence. The full corpus of 39 lectures (35.7 hours of audio) was transcribed in 43.7 minutes of wall-clock time, an average throughput of 49$\times$ realtime, consuming $\sim$4.3\,GB of VRAM. Individual lectures ranged from 37$\times$ to 56$\times$ realtime depending on the density of spoken content.

\paragraph{Domain vocabulary prompting.}
A significant challenge in transcribing technical lectures is the accurate recognition of domain-specific terminology. Out-of-the-box, Whisper frequently misrecognizes specialized terms that are rare in its general-purpose training data. In our initial transcription run, the word \emph{fugacity}---a central concept in chemical engineering thermodynamics---was consistently rendered as ``GASB.'' Other misrecognitions included ``Anaconda equation'' for the \emph{Antoine equation} and ``time-Robinson'' for the \emph{Peng-Robinson equation of state}.

Whisper's \texttt{initial\_prompt} parameter provides a mechanism to bias the decoder toward expected vocabulary. By supplying a prompt containing key thermodynamic terms (fugacity, Clausius-Clapeyron, Peng-Robinson, Poynting correction, etc.), we found that recognition of domain vocabulary improved substantially. With prompting, ``fugacity'' was correctly recognized throughout the transcript. Some terms (e.g., Peng-Robinson) remained occasionally garbled, motivating an expanded prompt including the full term ``Peng-Robinson equation of state'' rather than just ``Peng-Robinson.''

We additionally set \texttt{condition\_on\_previous\_text=False} in the transcription parameters. While this setting may reduce cross-segment coherence, it prevents the decoder from carrying misrecognitions or hallucinated context forward into subsequent audio windows.

\paragraph{Hallucination detection and mitigation.}

Whisper's autoregressive decoder is susceptible to a well-documented failure mode in which it enters a repetition loop, producing the same short phrase dozens of times in succession. In an initial baseline transcription of one lecture \emph{without} decoder-level suppression, we identified four such loops: ``Elizabeth.'' repeated 37 times (50.5\,s of audio), ``That's a lot.'' $\times 10$ (10.2\,s), ``You told me to.'' $\times 3$ (10.6\,s), and ``Okay.'' $\times 3$ (3.0\,s). In total, 53 of 826 raw segments (6.4\% by count, 2.2\% by time) were hallucinated repetitions.

These loops exhibit characteristic signatures: identical text across three or more consecutive segments, fabricated timestamps with exactly 1.0-second durations, and zero inter-segment gaps. The underlying audio during these episodes typically contains silence, student responses captured at low volume, or ambient classroom noise.

We address hallucination loops with a three-layer strategy:
\begin{enumerate}
\item \textbf{Decoder-level suppression}: A mild repetition penalty
  (\texttt{repetition\_penalty}$=1.1$) and exact 3-gram blocking
  (\texttt{no\_repeat\_ngram\_size}$=3$) in the CTranslate2 decoder.
\item \textbf{Context isolation}: Setting
  \texttt{condition\_on\_previous\_text=False} prevents a hallucinated
  segment from poisoning subsequent decoding windows.
\item \textbf{Post-processing}: A deterministic pass detects runs of three
  or more consecutive segments with identical text, retains the first
  occurrence, and removes the rest. Loop metadata (text, count, timestamps,
  duration) is recorded in the output JSON for quality auditing.
\end{enumerate}

Table~\ref{tab_transcript_quality} summarizes the effect of these mitigations
across the full lecture corpus. With all three layers active, the hallucination
rate dropped from 6.4\% of segments in the baseline to 0.02\% across the full
semester. Only a single loop was detected in 39 lectures (``What?'' $\times 5$
in lecture~30), and the post-processing filter removed 4 segments totaling
5\,seconds.

\begin{table}
\caption{\label{tab_transcript_quality}%
Transcript quality: baseline (one lecture, no suppression) versus full corpus
(39 lectures, all mitigations active).}
\begin{tblr}{hlines, colspec={lcc}}
\hline
\textbf{Metric} & \textbf{Baseline} & \textbf{Full corpus} \\
Lectures & 1 & 39 \\
Audio duration & 55.0 min & 35.7 hr \\
Transcription time & 177 s (18.6$\times$) & 43.7 min (49$\times$) \\
Raw segments & 826 & 20{,}361 \\
Hallucination loops & 4 & 1 \\
Segments removed & 53 (6.4\%) & 4 (0.02\%) \\
Time lost to hallucination & 74.2 s (2.2\%) & 5.0 s ($<$0.01\%) \\
Segments after cleaning & 773 & 20{,}357 \\
\hline
\end{tblr}
\end{table}

Of the 39 scheduled lecture sessions, 35 produced full transcripts (458--733 segments each), two sessions (010 and 025) were examinations with minimal proctor speech (2 and 5 segments, respectively), and two sessions (003 and 023) yielded no segments, likely due to class cancellations or silent audio.

\paragraph{The instructor as domain expert.}

The vocabulary prompting results illustrate a broader principle: the instructor's domain expertise is a critical input to the transcription pipeline. General-purpose speech recognition models, including those deployed with course recording software, are trained predominantly on conversational and broadcast speech, leaving them poorly calibrated for the specialized terms of any given academic discipline. The instructor, however, knows precisely which terms will appear in the lectures and can supply this knowledge to the system.

In our current implementation, this knowledge is encoded manually as a static vocabulary prompt. This approach is effective for a single thermodynamics course, but does not scale well. Each course or discipline will require a bespoke prompt, and terms may shift over the course of a semester as new topics are introduced. Several strategies could systematize the construction of domain-aware prompts:

\begin{itemize}
\item \textbf{Textbook index extraction.} The course textbook's table of contents and index provide a curated list of technical terms. These can be automatically parsed and used to generate per-chapter or per-topic vocabulary prompts that evolve alongside the course schedule.
\item \textbf{Syllabus-driven prompting.} The course syllabus maps topics to lecture dates. By aligning the vocabulary prompt to the syllabus, the transcription system can use a narrower, more targeted prompt for each lecture, reducing competition among unrelated terms in the decoder.
\item \textbf{Iterative refinement.} Misrecognized terms identified by the instructor during transcript review can be fed back into the prompt for subsequent transcription runs. Our pipeline stores all transcription parameters in the output JSON, supporting reproducible re-transcription as the prompt improves.
\end{itemize}

This feedback loop---in which the instructor's domain knowledge improves the quality of the data that ultimately feeds the learning assistant---is a distinctive feature of instructor-driven systems compared to fully automated approaches. The instructor is not merely a consumer of AI-generated transcripts but an active participant in tuning the underlying models to their specific pedagogical context. This reinforces the dual-facing design philosophy described in Section~\ref{sec:instructor_facing}. The same transcript pipeline that serves students as a searchable lecture archive simultaneously gives the instructor a tool for analyzing, refining, and improving their course materials.

\paragraph{Comparison with institutional transcripts}
\label{sec:kaltura_comparison}
To quantify the improvement from domain vocabulary prompting, we compared our Whisper-generated transcripts against the automatic speech recognition (ASR) transcripts provided by the university's Kaltura lecture-capture system. Six matched lecture pairs (lectures~26--31) were available in both formats, matched by a shared recording identifier in the filename.

At the aggregate level, the two systems are remarkably similar. Table~\ref{tab_kaltura_whisper} summarizes the comparison. Word counts are within 1\% for all six lectures. High-frequency domain terms---\emph{entropy}, \emph{enthalpy}, \emph{adiabatic}, \emph{reversible}, \emph{thermodynamic}---appear at nearly identical rates in both transcripts, indicating that both ASR systems have sufficient exposure to common scientific vocabulary in their training data.
\begin{table}
\caption{\label{tab_kaltura_whisper}%
Comparison of institutional (Kaltura) and domain-prompted Whisper transcripts for six matched lectures. Domain term counts are summed across all six lectures.}
\begin{tblr}{hlines, colspec={lcc}}
\hline
\textbf{Metric} & \textbf{Kaltura} & \textbf{Whisper} \\
Mean word count & 7,449 & 7,409 \\
Word count ratio & \SetCell[c=2]{c} 0.99--1.01 & \\
``entropy'' (total) & 329 & 331 \\
``enthalpy'' (total) & 50 & 49 \\
``reversible'' (total) & 87 & 81 \\
``thermodynamic(s)'' (total) & 117 & 115 \\
\hline
\end{tblr}
\end{table}
The real difference lies in \emph{contextual fidelity}---the accuracy of words surrounding the domain terms and the recognition of less common scientific vocabulary. Table~\ref{tab_kaltura_examples} presents representative passage-level differences from lecture~9 (introduction to entropy). In each case, Kaltura produces a phonetically plausible but semantically nonsensical substitution, while the domain-prompted Whisper transcript preserves the intended meaning.
Notably, both systems fail on rare proper nouns: \emph{Helmholtz} is rendered as ``compulse'' by both Kaltura and Whisper. This suggests that even domain-prompted models struggle with terms that are uncommon in general speech corpora; further vocabulary expansion or fine-tuning may be needed for such cases.

The Kaltura JSON format provides per-word confidence scores, enabling a post-hoc check. Garbled words consistently show low confidence (0.32--0.67 for the examples above), while correctly recognized domain terms score 0.88--1.00. In principle, low-confidence words could be flagged for correction, but the Kaltura plain-text export strips this metadata entirely.

\begin{table}
\caption{\label{tab_kaltura_examples}%
Representative misrecognitions comparing Kaltura and domain-prompted
Whisper transcripts (lecture~9). The intended phrase is determined
from surrounding context.}
\begin{tblr}{hlines, colspec={Q[p,wd=0.22\textwidth]Q[p,wd=0.22\textwidth]Q[p,wd=0.22\textwidth]}}
\hline
\textbf{Intended} & \textbf{Kaltura} & \textbf{Whisper} \\
the next module & the next \emph{logro} & the next module \\
liquid versus gases & liquid versus \emph{gaffy} & liquids versus gases \\
a balance of the enthalpies & a balance of the \emph{envelopes of the trees} & enthalpies \\
Gibbs and Helmholtz free energies & \emph{ideas} and \emph{compulse} free energies & \emph{gives} and \emph{compulse} for energies \\
keeping your protons & keeping your \emph{protein} & keeping your protons \\
keeping your quarks & keeping your \emph{torch} & keeping your quarks \\
thermodynamics tells us & thermodynamics \emph{Electronics} tells us & thermodynamics tells us \\
\hline
\end{tblr}
\end{table}

Beyond recognition accuracy, the two formats differ structurally. Kaltura produces a wall of continuous text with no segmentation or timestamps. Whisper produces sentence-level segments, each with start and end timestamps available in a companion JSON file. This structural difference is consequential for downstream processing: our lecture analysis pipeline (Section~\ref{sec:lecture_analysis}) relies on timestamped segments for question identification and confusion detection, analyses that cannot be performed on unsegmented text.

A practical barrier to deploying the lecture analysis pipeline on a per-lecture cadence during a semester is the difficulty of obtaining lecture recordings from institutional systems. The university's Kaltura-based lecture capture system provides no bulk download API; recordings must be accessed one at a time through multiple manual steps. Even for institutional administrators, the process is equally manual. This vendor lock-in means that while the technical pipeline can process a
lecture in minutes, retrieving the recording remains a significant bottleneck. Batch processing at the end of a semester is currently the most practical approach.

A simpler alternative that bypasses the institutional system entirely is direct audio capture using a USB microphone worn by the instructor. A lapel or headset microphone connected to a laptop or phone produces a high-quality, close-capture recording that can be processed immediately after class. This approach eliminates the Kaltura download barrier, avoids vendor lock-in, provides better audio quality than a room-mounted capture microphone (reducing ASR errors), and enables a per-lecture workflow in which the summary of one lecture is available to students before the next class meeting.

Real-time, per-lecture analysis---where a structured summary is distributed to students as a spaced retrieval cue before the next class---is technically ready but deployment-blocked by the institutional recording workflow. We advocate for open API access to institutional lecture recordings as a prerequisite for these tools to reach their full potential. In the meantime, direct capture with a USB microphone represents a practical and effective workaround.


\section{Student-facing tool: textbook query}
\label{sec:rag}

The student-facing query pipeline uses retrieval-augmented generation to ground LLM responses in the course textbook. \cite{sandler2017chemical} Rather than relying on the model's parametric knowledge---which risks plausible but incorrect thermodynamic statements---the system retrieves specific textbook sections relevant to the student's question and constrains the LLM to synthesize a response from that evidence alone.

\subsection{Dual-path term extraction}

A naive approach to matching student queries against the textbook index would be to compute embedding-based similarity between the query and index entries. We chose instead a term-extraction approach: identify the technical concepts in the query, then search the index for entries containing those terms. This is more transparent (students and instructors can see \emph{why} a match was returned), more lightweight (no embedding model or vector store required), and well-suited to the structured nature of a back-of-book index.

To make term extraction robust, we run two extractors in parallel (Figure \ref{fig:pipeline}):
\begin{enumerate}
\item \textbf{Regex extraction.} A pattern-based extractor that removes stopwords, recognizes predefined multi-word technical phrases (``van~der~Waals,'' ``equation~of~state,'' ``chemical~potential,'' etc.), and returns both the recognized phrases and individual content words. This extractor is fast, deterministic, and requires no model inference, but is limited to its predefined vocabulary.
\item \textbf{LLM extraction.} A local LLM (Llama~3.1 8B via Ollama) receives the student query with a system prompt instructing it to extract search terms and suggest related concepts. For example, the query ``What is entropy?'' yields not only \emph{entropy} but also \emph{entropy change}, \emph{entropy generation}, and \emph{entropy balance}---query expansion that the regex extractor cannot perform. If the LLM is unavailable, the system falls back gracefully to regex-only extraction.
\end{enumerate}

Both extractors feed their terms into the same index search function, which walks the hierarchical index recursively, scoring each entry based on term overlap weighted by term length (longer, more specific terms score higher), with bonuses for matches at the start of a topic name. The LLM path additionally incorporates a page-position bonus, giving higher weight to earlier pages where fundamental concepts are introduced.

The dual-path extraction addresses a practical fragility of local LLM inference: small models are not perfectly reliable on every query. By running a deterministic regex extractor in parallel, the system guarantees a baseline of results even when the LLM produces unexpected output. The max-score merge ensures that the better of the two methods dominates for each match.

\subsection{Max-score merge}

The results from the two extractors are merged using a max-score strategy. For each unique index entry (identified by its topic and page number), the system takes the \emph{maximum} score from either method. This is deliberately asymmetric. If the regex extractor finds a relevant entry that the LLM misses (or vice versa), the entry is preserved at full score rather than being penalized. An earlier weighted-average approach (0.7$\times$LLM + 0.3$\times$regex) was abandoned because it suppressed valid matches that only one extractor found---for example, ``fugacity coefficient'' matched by regex but missed by the LLM on a given run would receive a blended score of only 6.3 (below the threshold) instead of its true score of 21.

After merging, results below a score threshold are filtered, entries without page numbers are excluded, and the remaining matches are sorted by page number (ascending) as the primary key, with score as a tiebreaker. This page-ordered presentation prioritizes foundational treatments of a concept over specialized applications that appear later in the textbook.

\subsection{Context assembly and LLM synthesis}

The top five merged results are assembled into a context string for the answer-generation LLM. Each result is enriched with chapter and section information from the table-of-contents navigation tree, producing entries such as:
\begin{quote}
\small
\texttt{1. Entropy > Entropy generation (pages 958--962)} \\
\texttt{~~~Location: Chapter~15, Section~15.7 (Thermodynamic Analysis of Bioreactors)}
\end{quote}

This context, together with the student's original query, is passed to the LLM (Llama~3.1 8B, temperature 0.6) with a system prompt that instructs it to act as a helpful teaching assistant. The prompt includes explicit grounding constraints: the model must use \emph{only} the provided context, must not invent chapter or section numbers, and must include specific page references for every topic mentioned. The output is a concise (3--5 sentence) natural-language response that addresses the student's question and directs them to the relevant textbook sections.

If the LLM is unavailable, a deterministic fallback produces a simpler response using the top-scoring match (``Check out \emph{[topic]} on pages~[pages]''). This ensures the system remains functional even when the model server is down.

\subsection{Example query}

To illustrate the full pipeline, consider a student query: ``Explain fugacity.'' The two extractors operate in parallel:
\begin{itemize}
\item \textbf{Regex extractor.} After stopword removal, extracts the single
  term \emph{fugacity}.
\item \textbf{LLM extractor.} Returns \emph{fugacity}, \emph{fugacity
  coefficient}, \emph{chemical potential}, \emph{activity coefficient},
  \emph{ideal gas}, and \emph{non-ideal gas}---expanding the query to
  related concepts that the regex path cannot infer.
\end{itemize}
Both term sets are searched against the index. The max-score merge produces a ranked list; after filtering and page-sorting, the top results sent to the answer-generation LLM are:
\begin{quote}
\small
\begin{enumerate}[noitemsep, topsep=0pt]
\item Peng-Robinson equation of state $>$ fugacity coefficient from (pp.~314--317, 440--442) \\
  \emph{Chapter~7, Section~7.4 (The Molar Gibbs Energy and Fugacity of a Pure Component)}
\item Corresponding states $>$ fugacity coefficient (pp.~315--316) \\
  \emph{Chapter~7, Section~7.4}
\item Liquid(s) $>$ fugacity of (pp.~316--317) \\
  \emph{Chapter~7, Section~7.4}
\item Solid(s) $>$ fugacity of (p.~320) \\
  \emph{Chapter~7, Section~7.4}
\item Phase equilibrium $>$ fugacity in (pp.~424--425) \\
  \emph{Chapter~9, Section~9.2 (The Partial Molar Gibbs Energy and Fugacity)}
\end{enumerate}
\end{quote}

The LLM synthesizes these results into the student-facing response shown in Figure~\ref{fig:example_query}. Three mechanisms work together to keep the response grounded: (1)~\emph{architectural}---the model receives only the five retrieved context entries, never the full index or textbook text, so unrelated material is simply unavailable;(2)~\emph{prompt-level}---the system prompt explicitly instructs the model to use only the provided context and to never fabricate chapter or section numbers; and (3)~\emph{temperature}---inference at $T=0.6$ balances natural, conversational phrasing against the higher hallucination risk of more creative sampling. Together, these constraints make it unlikely (though not impossible) for the model to introduce information that is not present in the retrieved evidence.
\begin{figure}[h]
\fbox{\begin{minipage}{0.92\columnwidth}
\raggedright
\small
\textbf{Student query:} ``Explain fugacity.''
\medskip

\textbf{Generated response (Llama~3.1 8B):}
\smallskip
Fugacity is a measure of the tendency of a component to escape from a
system or to be driven out of equilibrium. According to the textbook,
fugacity is related to the molar Gibbs energy and can be calculated using
the equation mentioned in Chapter~7, Section~7.4 (The Molar Gibbs Energy
and Fugacity of a Pure Component).

The Peng-Robinson equation of state is also relevant here, as it provides
a way to calculate the fugacity coefficient from the given parameters. To
understand how fugacity behaves in different phases, such as liquids and
solids, we can refer to Chapter~7, Section~7.4. When dealing with phase
equilibrium, it's essential to consider the role of fugacity in
determining the equilibrium state. This is covered in Chapter~9,
Section~9.2 (The Partial Molar Gibbs Energy and Fugacity).

\smallskip
\emph{References:}
Peng-Robinson equation of state: pages 314--317, 440--442 in Chapter~7,
Section~7.4.
Corresponding states: pages 315--316 in Chapter~7, Section~7.4.
Liquid(s): pages 316--317 in Chapter~7, Section~7.4.
Solid(s): page~320 in Chapter~7, Section~7.4.
Phase equilibrium: pages 424--425 in Chapter~9, Section~9.2.
\end{minipage}}
\caption{\label{fig:example_query}%
Example query and generated response from the RAG pipeline. The response
is produced by Llama~3.1 8B at temperature $T=0.6$, constrained to use
only the five retrieved index entries. All chapter numbers, section
titles, and page references in the output are traceable to the retrieved
context.}
\end{figure}

\section{Instructor-facing tool: lecture analysis}
\label{sec:instructor_tool}

The same transcript corpus that supports student queries also enables structured content analysis for the instructor. Using local LLM inference, the system processes each lecture transcript through four analysis types---\emph{summary}, \emph{questions}, \emph{confusion detection}, and \emph{anecdotes/analogies}---producing structured JSON output that supports course reflection and improvement.

All analyses used the Llama~3.1 8B model via Ollama with a 16{,}384-token context window (sufficient for full lecture transcripts of $\sim$13{,}000 tokens) and JSON-mode output. The full corpus of 35 lectures was processed across all four analysis types in approximately 15~minutes on the RTX~4090 workstation.

\subsection{Analysis pipeline}
\label{sec:lecture_analysis}

\paragraph{Lecture summaries.}
For each lecture, the model produces a structured summary containing a title, a list of topics with descriptions, key concepts, key equations, a narrative summary, and a lecture-type classification (new material, review, problem solving, etc.). All 35 lectures produced schema-compliant output, with content that verifiably matches the transcripts---topic identifications align with textbook chapters, equation references correspond to derivations in the audio, and lecture-type classifications are consistent with the course schedule.

These summaries serve both instructors and students. For the instructor, they provide a semester-at-a-glance overview useful for course planning and identifying gaps in topic coverage. For students, summaries distributed between lectures can function as spaced retrieval cues at the edge of forgetting---priming concepts from the previous lecture right before students need to build on them in the next class. Research on the spacing effect suggests that such timely reminders are more effective for long-term retention than massed review. \cite{roediger2011critical,karpicke2008critical}

Figure~\ref{fig:example_summary} shows a representative summary for lecture~9 (introduction to entropy), the same lecture used in the Kaltura comparison (Section~\ref{sec:kaltura_comparison}). The model correctly identifies the lecture's four main topics, extracts relevant concepts, and classifies it as new material---all from the raw transcript in 5.6~seconds of inference time.
\begin{figure}[h]
\fbox{\begin{minipage}{0.92\columnwidth}
\raggedright
\small
\textbf{Lecture 9 --- Summary} (Llama~3.1 8B, 5.6\,s)
\medskip

\textbf{Title:} Entropy and Direction of Thermodynamic Processes \\
\textbf{Type:} New material
\smallskip

\textbf{Topics:}
\begin{enumerate}[noitemsep, topsep=2pt, leftmargin=*]
\item \emph{Introduction to Entropy.} Definition of entropy as disorder
  or randomness, and its relation to thermodynamic processes.
\item \emph{Spontaneous Processes in Isolated Systems.} How spontaneous
  processes proceed towards equilibrium.
\item \emph{Directionality of Thermodynamic Processes.} Why energy is
  not always accessible to us, and how this relates to directionality.
\item \emph{Entropy as a State Function.} Entropy as a new state
  function that determines the direction of a process.
\end{enumerate}

\smallskip
\textbf{Key concepts:} entropy, directionality of thermodynamic
processes, spontaneous processes, equilibrium state

\smallskip
\textbf{Summary:} This lecture introduced the concept of entropy as a
measure of disorder or randomness in a system, and its relation to the
directionality of thermodynamic processes. The instructor discussed how
spontaneous processes in isolated systems proceed towards equilibrium,
and how energy is not always accessible to us. The concept of entropy was
introduced as a new state function that determines the direction of a
process.
\end{minipage}}
\caption{\label{fig:example_summary}%
Structured summary produced by the lecture analysis pipeline for
lecture~9 (introduction to entropy). All fields are extracted
automatically from the raw transcript by Llama~3.1 8B in JSON mode.
The full JSON output includes additional fields (key equations, source
file, model metadata) omitted here for brevity.}
\end{figure}

\paragraph{Question identification.}
The pipeline identifies questions from the timestamped transcript, classifying each by speaker (student or instructor), type (conceptual, clarification, procedural, or Socratic), and pedagogical significance (high, medium, or low). This analysis proved the most challenging for the local model; we discuss the failure modes and our mitigation strategy in Section~\ref{sec:failure_modes}.

Figure~\ref{fig:example_questions} shows representative output from the same entropy lecture (lecture~9).  The model captures instructor Socratic prompts (``How do we know which direction this process should run?'') and assigns relevance ratings that reflect pedagogical significance. Some instructor prompts, e.g.\ ``What is entropy?'' are mislabled as student questions. The results are presented verbatim from the tool and include an example of a poorer result. For instance, question 5 in Figure~\ref{fig:example_questions} is a confusing statement about the equivalence of work and heat energy. In the class, students learn to recognize their equivalence in terms of the conservation of energy, and their difference in terms of the quality of energy represented by the entropy concepts introduced in the lecture. Similar minor quality issues appear in all of the semantic tools (questions, confusion detection, anecdotes and analogies).

\begin{figure}[h]
\fbox{\begin{minipage}{0.92\columnwidth}
\raggedright
\small
\textbf{Lecture 9 --- Questions} (Llama~3.1 8B, two-pass)
\medskip

\begin{enumerate}[noitemsep, topsep=2pt, leftmargin=*]
\item \texttt{[0:00:20]} (S) \textbf{[high]} What is entropy?
\item \texttt{[0:04:03]} (I) \textbf{[high]} Can entropy be reduced?
\item \texttt{[0:34:00]} (I) \textbf{[high]} How do we know which direction
  this process should run?
\item \texttt{[0:41:42]} (S) \textbf{[high]} So it's like if you want to
  remove all those gradients?
\item \texttt{[0:45:04]} (I) \textbf{[high]} What do I mean by work is
  not\ldots\ Work energy is equivalent to heat energy, but work is not.
\end{enumerate}

\smallskip
\textit{11 questions identified (6 student, 5 instructor)}
\end{minipage}}
\caption{\label{fig:example_questions}%
Questions extracted from lecture~9 (introduction to entropy) by the two-pass pipeline.  Each entry includes a timestamp, speaker tag (S\,=\,student, I\,=\,instructor), and relevance rating.  Five of eleven identified questions are shown.}
\end{figure}

\paragraph{Confusion detection.}
Using the timestamped transcript, the model identifies moments of apparent student confusion or instructor re-explanation, including the topic, evidence (e.g., ``instructor restarts explanation from a different angle''), and a severity rating (minor, moderate, or significant). All 35~lectures produced results, though the analysis tends toward over-sensitivity: normal pedagogical rephrasing is sometimes flagged as confusion.

Figure~\ref{fig:example_confusion} shows confusion points detected in the entropy lecture.  The model identifies specific conceptual difficulties and assigns severity levels, providing the instructor with a timestamped map of where students struggled. The first example (``Entropy and its relation to disorder or randomness'') is a particularly important reminder for a thermodynamics instructor to discuss with students their pre-conceived notions of entropy in contrast to its introduction as a quantitative state function. 

\begin{figure}[h]
\fbox{\begin{minipage}{0.92\columnwidth}
\raggedright
\small
\textbf{Lecture 9 --- Confusion Detection} (Llama~3.1 8B)
\medskip

\begin{enumerate}[noitemsep, topsep=2pt, leftmargin=*]
\item \texttt{[0:00:37]} \textbf{[minor]} \emph{Entropy and its relation to disorder or randomness.} Students seem unfamiliar with the concept of entropy, requiring an explanation from the instructor.
\item \texttt{[0:15:42]} \textbf{[moderate]} \emph{Directionality of thermodynamic processes and the role of entropy.} Students seem unsure about how to determine the direction of a thermodynamic process using entropy.
\item \texttt{[0:45:04]} \textbf{[moderate]} \emph{Relationship between work and heat energy.} Students may struggle to understand the distinction between work and heat energy in thermodynamics.
\item \texttt{[0:50:08]} \textbf{[moderate]} \emph{Internal energy and its accessibility as useful work.} Students may be unclear on how to determine which forms of internal energy are accessible for use in a thermodynamic system.
\end{enumerate}

\smallskip
\textit{6 confusion points identified in this lecture}
\end{minipage}}
\caption{\label{fig:example_confusion}%
Confusion points detected in lecture~9 (introduction to entropy).  Each entry includes a timestamp, severity level, topic, and evidence description.  Four of six detected points are shown.}
\end{figure}

\paragraph{Anecdotes and analogies.}
The model catalogs the instructor's use of anecdotes, analogies, jokes, real-world examples, demonstrations, and historical notes, capturing a verbatim quote, a description, the related topic, and the pedagogical purpose of each item. All 35~lectures produced results with good variety across categories. This analysis is primarily instructor-facing, supporting reflection on teaching style and the identification of recurring explanatory strategies.

Figure~\ref{fig:example_anecdotes} illustrates the categorized output from the entropy lecture.  The model captures a range of pedagogical devices---from personal anecdotes and historical references to real-world demonstrations---each with a verbatim quote from the transcript. We reproduce the script output verbatim, which captures some errors, including ``Prof.\ Wang's definition of entropy'' and the cooling of an isolated system. 

\begin{figure}[h]
\fbox{\begin{minipage}{0.92\columnwidth}
\raggedright
\small
\textbf{Lecture 9 --- Anecdotes \& Analogies} (Llama~3.1 8B)
\medskip

\begin{enumerate}[noitemsep, topsep=2pt, leftmargin=*]
\item \textbf{[anecdote]} Professor Wang's definition of life. \\
  \emph{``The definition of life is something that keeps energy and reduces
  its entropy.''}
\item \textbf{[real\_world\_example]} Thermal equilibration of metal cubes. \\
  \emph{``If I put two cubes of metal at different temperatures in an
  insulated box, they will thermally cool it, and they will not
  spontaneously un-cool it.''}
\item \textbf{[historical\_note]} Joule's experiment on work--heat
  equivalence. \\
  \emph{``Joule shows us this in his experiment\ldots\ work energy is
  equivalent to heat energy.''}
\item \textbf{[analogy]} Limits of heat-to-work conversion. \\
  \emph{``But heat cannot be converted entirely to work.  This conversion is
  possible but never 100\% efficient.''}
\end{enumerate}

\smallskip
\textit{7 items identified (1 anecdote, 2 analogies, 1 real-world example,
1 historical note, 1 joke, 1 story)}
\end{minipage}}
\caption{\label{fig:example_anecdotes}%
Anecdotes and analogies cataloged from lecture~9 (introduction to
entropy).  Each item is categorized by type and includes a verbatim
quote from the transcript.  Four of seven items are shown.}
\end{figure}

\subsection{Failure modes and mitigations}
\label{sec:failure_modes}

Running structured extraction tasks on an 8B-parameter local model over a corpus of 35 lectures revealed several systematic failure modes. We present these as a practical taxonomy for practitioners building similar LLM-over-transcript pipelines.

\paragraph{Context truncation.}
Ollama's default context window is 2{,}048 tokens, but lecture transcripts are approximately 13{,}000 tokens. When the context window parameter (\texttt{num\_ctx}) was not explicitly set, the model received only the first $\sim$15\% of each transcript. The resulting outputs were superficially well-formed---valid JSON with plausible-sounding content---but bore little relation to the actual lecture. Failure presentations included empty result objects, plain-text narratives instead of structured schemas, and hallucinated data (fabricated exam problems, fictional Q\&A datasets with \texttt{example.com} URLs). Crucially, these failures were \emph{silent}: the model did not indicate that input had been truncated. Analysis times were anomalously fast ($\sim$2\,s versus $\sim$60\,s with full context), providing the only external signal of the problem. Setting \texttt{num\_ctx}$= 16{,}384$ resolved all truncation failures.

\paragraph{Placeholder echoing.}
When the system prompt included a schema example with placeholder values (e.g., \texttt{"timestamp": "H:MM:SS"}), the model sometimes copied the placeholder literally instead of extracting actual values from the transcript. In our confusion-detection analysis, all 35~lectures initially showed the literal string \texttt{H:MM:SS} for every timestamp. Replacing the placeholder with a concrete example (\texttt{"0:15:42"}) and adding an explicit instruction (``Do NOT use placeholder text'') resolved the issue.

\paragraph{Over-classification.}
The model consistently identified more items than warranted. In question extraction, filler phrases (``right?'', ``okay?'') and classroom management utterances (``Any questions?'') were classified as substantive questions. In anecdote detection, routine worked examples were tagged as ``anecdotes.'' In confusion detection, normal pedagogical rephrasing was flagged as evidence of student difficulty. While the raw over-extraction is noisy, it biases toward recall over precision---a more useful failure mode for downstream filtering than under-extraction, which would silently lose information.

\paragraph{Schema drift.}
Despite requesting a specific JSON schema, the model occasionally used different field names (\texttt{topic} versus  \texttt{name}, \texttt{notes} versus \texttt{description}), included unexpected fields, or inserted bare strings into arrays that should contain objects. Ollama's JSON mode guarantees syntactically valid JSON but does not enforce schema compliance. Defensive parsing in the report formatter (type-checking array elements, falling back to alternative field names) was necessary to handle the variation.

\paragraph{Bimodal output in question extraction.}
The most striking failure mode appeared in question identification. With a one-shot prompt that asked the model to simultaneously extract, classify, and filter questions, the output distribution was sharply bimodal: 26 of 35 lectures produced exactly 8 questions each (with fabricated timestamps, uniform classification, and evenly spaced entries), while 5 lectures produced 64--108 questions (exhaustively extracting every conversational exchange). The ``exactly 8'' mode indicates the model defaulting to a fixed output length and generating formulaic study-guide-style content rather than analyzing the transcript. The high-count mode represents the opposite failure: no filtering at all.

This bimodal behavior suggests that an 8B model cannot reliably perform extraction, classification, and filtering as a single composite task over a long input. We addressed this with a \emph{two-pass architecture}: Pass~1 (extraction) reads the full transcript and identifies candidate questions with verbatim quotes---a simple, high-recall task grounded in the source text. Pass~2 (filtering) receives only the candidate list ($\sim$1--2{,}K tokens instead of $\sim$13{,}K) and classifies, ranks, and filters to the 5--15 most significant questions. With this decomposition, the output distribution became smooth (range 2--15, mean 8.8, no clustering), candidate counts varied naturally across lectures (2--46), and the filter pass demonstrably reduced over-extracted lectures (e.g., 46~candidates, 11~kept). The two-pass approach doubled per-lecture processing time but eliminated the bimodal failure entirely.

\paragraph{Repetitive output.}
In several lectures the confusion-detection analysis produced dozens of near-identical entries for the same topic.  For example, one lecture on table interpolation generated 28 consecutive confusion entries all reading ``Interpolation of thermodynamic properties from tables: The instructor is having trouble with the correct sign for interpolation.''  This is distinct from over-classification: the model correctly identifies the topic but fails to consolidate repeated occurrences into a single entry, apparently treating each segment of the transcript independently.  Post-processing deduplication (grouping entries by topic and timestamp proximity) is a straightforward mitigation that we have not yet implemented.

\section{Python Library Implementation}

Stan is implemented as a modular Python library, structured to support both ease of use and extensibility. \cite{GHrepo} The system architecture is organized into a primary \texttt{stan} package containing the core LLM logic, complemented by an \texttt{apps} directory for end-user interfaces. To promote transparency and community-driven development, the repository follows a standard distribution format:
\begin{itemize}
    \item \textbf{Core Logic.} The \texttt{stan/} directory encapsulates the backend data processing, 
    prompt engineering, and model management.
    \item \textbf{Documentation and Reproducibility.} Comprehensive guides are provided in \texttt{docs/} 
    writen with python's \texttt{sphinx} library, \cite{sphinx2026} while the \texttt{examples/} 
    directory offers practical templates in the form of Jupyter Notebooks. \cite{kluyver2016jupyter}
    \item \textbf{Configuration.} A \texttt{pyproject.toml} file ensures consistent dependency management 
    and cross-platform compatibility, facilitating local deployment on various hardware configurations.
\end{itemize}
This structure ensures that the tool is not merely a static application but a flexible library that can be imported and adapted for specific research and teaching requirements.  The modular design specifically enables the integration of interactive computational modules--such as thermodynamic property solvers, Equations of State (EOS) calculators, phase equilibria solvers--allowing the LLM assistant to serve as a high-level interface for executing and sharing examples of these engineering routines with students.

\section{Future Work}

We envision the future development of Stan proceeding along three interrelated trajectories: knowledge expansion, interactive computational integration, and curricular scaling.

\textbf{Knowledge Expansion and Tool Integration.}
Planned iterations will expand the retrieval index to include foundational chemical engineering principles, governing equations, and physical explanations drawn directly from core textbooks. We also aim to strengthen connections to contemporary literature to provide instructors with updated case studies and examples.

A primary technical objective is the integration of transparent, pedagogical computational tools. Current development includes:
\begin{itemize}
    \item Porting original MATLAB-based routines for the Peng--Robinson Equation of State (EOS) and UNIFAC models \cite{sandler2017chemical} into Python.
    \item Vapor--Liquid Equilibrium (VLE) solvers for flash calculations.
    \item Unit conversion utilities for Henry's Law constants.
    \item Pedagogical implementations of nested bisection and \texttt{scipy}-based root-finding routines for phase equilibria.
\end{itemize}
These scripts are intentionally written for conceptual clarity rather than computational optimization. Commercial platforms such as Aspen Plus, MATLAB, and gPROMS remain standard in research and industrial practice. The Python-based implementations instead function as ``glass box'' models that promote computational literacy and enable the LLM to reference concrete, inspectable examples when responding to student queries.

\textbf{Curricular and Institutional Scaling.}
Scaling beyond a single course requires administrative and structural development. Planned features include mechanisms for streamlined course handovers between instructors and continuity of indexed data across semesters. 

A broader challenge involves extending the hierarchical indexing framework across the chemical engineering curriculum. Differences in terminology and notation, and course-specific context must be reconciled without flattening disciplinary nuance---or perhaps our developing tools can help students understand and incorporate those differences into their engineering learning and practice. Addressing these issues will require coordinated departmental engagement rather than isolated implementation. 

Finally, systematic evaluation remains essential. Future work will include defining measurable learning outcomes, assessing cognitive engagement, and identifying metrics that allow iterative refinement of the system. Establishing rigorous assessment frameworks will determine whether tools such as Stan substantively improve student learning trajectories within technical disciplines.

\section{Conclusion}

This work introduced Stan, a locally deployable, LLM-based learning thermodynamics assistant designed for user-driven adaptation within the chemical engineering curriculum. Stan implements both student-facing and instructor-facing tools, the latter of which remains comparatively underrepresented in discussions of AI in the classroom. \cite{wang2025} Rather than automating core instructional labor such as grading or content generation, \cite{sridhar2023,reza2025} Stan is designed to extend instructional capacity. Its instructor-facing tools are primarily analytic. They examine and structure what has already occurred in the classroom, enabling reflection, pattern recognition, and iterative course refinement.

A central architectural feature of Stan is local deployment. By running on existing student and faculty hardware, Stan addresses issues of cost predictability, data governance, and privacy while avoiding usage-based pricing models. Local inference effectively converts existing hardware---already provisioned for academic work---into productive edge computing AI capacity. In alignment with open-source principles, the system architecture as a standard Python package is designed to encourage inspection, modification, and extension by instructors and students. With the emergence of fully open models such as Apertus, \cite{apertus2025} it is now feasible to construct an entirely open educational AI stack without proprietary dependencies. 

This distributed strategy, however, introduces important constraints. Hardware heterogeneity across student devices limits model size and performance. Quantized 7B-class models are broadly accessible, whereas larger models may exclude a subset of students. Performance variability may introduce pedagogical inequities even if financial barriers are avoided. In addition, shifting from cloud-based services to local infrastructure could transfer costs from subscription fees to instructional support time, including installation, dependency management, and cross-platform troubleshooting by instructors and teaching assistants.

Beyond infrastructure, the deployment of Stan raises broader pedagogical questions. Will the availability of high-fidelity transcription and timestamped query systems influence student attention patterns during live lectures? The integration of LLM-based assistants may also encourage more explicit verbalization by instructors, as deictic expressions (e.g., ``this term'') lose clarity in transcription. These shifts suggest that generative AI tools may not merely assist existing practices but could subtly reshape communicative norms within technical instruction.

Stan is therefore best understood not as a replacement for instructional practice, but as a catalyst for re-examining how knowledge is documented, analyzed, and extended in engineering education. Its development over the past year suggests that locally deployable LLM systems can be integrated into technically rigorous curricula like thermodynamics while preserving instructor agency and institutional autonomy.

The foundation of our thermodynamics course and Stan is a carefully curated thermodynamics textbook. \cite{sandler2017chemical} The value of such a resource---both pedagogically in the class and as a grounding corpus for LLM-assisted instruction---cannot be overstated. Recent work has shown that training on textbook-quality data enables small models to match much larger ones, \cite{li2023a} and that even state-of-the-art LLMs struggle with unsupervised thermodynamics reasoning without domain grounding. \cite{geissler2025} Rather than replace the course book, Stan extends it, making its content more readily accessible to students through a natural language interface and tighter integration with lectures.

\section*{Acknowledgments}
We would like to thank our University of Delaware Department of Chemical and Biomolecular Engineering students, especially those from the Fall 2025 class, for their engagement and feedback throughout the course. E.M.F.\ would additionally like to thank R.\ Lobo and E.\ Slaughter, his co-instructors, teaching assistants S.\ Meil, H.\ Zucco, and M. Pitell for input into the course content, and B.\ Kirby for fruitful discussions on pedagogical applications of LLMs in engineering courses.  
V.V.\ would like to thank A.\ N.\ Beris and J.\ L.\ Plawsky for their guidance and several fruitful discussions over the 
years on thermodynamics and pedagogy.


%

\end{document}